# Convolutional Neural Networks for Sentiment Analysis on Weibo Data: A Natural Language Processing Approach

1st Yufei Xie [1, 2]

[1] College Of Computing And Information Technologies

National University

Manila, Philippines

[2] School of Information Engineering

Jiangxi College of Applied Technology

Ganzhou, China

xiey@students.national-u.edu.ph

2nd Rodolfo C. Raga Jr.

College of Computer and Information Technologies

National University

Manila, Philippines

rjrcraga@national-u.edu.ph

*Abstract*—This study addressed the complex task of sentiment analysis on a dataset of 119,988 original tweets from Weibo using a Convolutional Neural Network (CNN), offering a new approach to Natural Language Processing (NLP). The data, sourced from Baidu's PaddlePaddle AI platform, were meticulously preprocessed, tokenized, and categorized based on sentiment labels. A CNN-based model was utilized, leveraging word embeddings for feature extraction, and trained to perform sentiment classification. The model achieved a macro-average F1-score of approximately 0.73 on the test set, showing balanced performance across positive, neutral, and negative sentiments. The findings underscore the effectiveness of CNNs for sentiment analysis tasks, with implications for practical applications in social media analysis, market research, and policy studies. The complete experimental content and code have been made publicly available on the Kaggle data platform for further research and development. Future work may involve exploring different architectures, such as Recurrent Neural Networks (RNN) or transformers, or using more complex pre-trained models like BERT, to further improve the model's ability to understand linguistic nuances and context.

## I. INTRODUCTION

Sentiment analysis, also known as opinion mining, has become an indispensable tool in the age of digital media where public opinion can be swiftly gauged from a plethora of social media platforms. This computational study of people's emotions, attitudes, and opinions has far-reaching applications ranging from brand management and market research to policy-making and political forecasting.

One such fertile ground for sentiment analysis is Weibo, a popular microblogging platform in China that boasts over 500 million active users. This platform offers a wealth of user-generated content, encompassing a myriad of topics and, consequently, a spectrum of sentiments, making it an ideal data source for sentiment analysis.

However, the manual evaluation of sentiment is arduous and practically unscalable given the immense volume of data. This has necessitated the application of machine learning (ML) techniques for automatic sentiment analysis. Among various ML approaches, Convolutional Neural Networks (CNNs) have proven to be particularly effective. CNNs, originally designed for image processing, have demonstrated their prowess in Natural Language Processing (NLP) tasks such as sentiment analysis due to their ability to capture local dependencies in the input data, a key factor when working with text.

CNNs employ a hierarchical layer structure to learn increasingly complex features from raw input, enabling them to capture intricate linguistic structures that could be crucial for sentiment analysis. CNNs' ability to efficiently deal with high-dimensional data, coupled with their lower need for pre-processing compared to traditional NLP techniques, makes them an attractive choice for sentiment analysis on Weibo data.

In this paper, we delve into the application of CNNs for sentiment analysis on Weibo data, aiming to explore the efficacy of this approach in the realm of NLP tasks and assess its potential implications. The complete experimental content and code have been made publicly available on the Kaggle data platform for further research and development[1].

## II. LITERATURE REVIEW

Sentiment analysis has been widely used to understand public attitudes and opinions in various domains, such as public opinion analysis, child abuse attitudes, policy sentiment analysis, and more. With the advent of machine learning and deep learning techniques, sentiment analysis has gained more precision and efficiency. This review aims to encapsulate the evolution of sentiment analysis approaches with a particular focus on those applied to Weibo data and highlight the potential of Convolutional Neural Networks (CNNs) in this field.

Li et al. (2022) proposed a sentiment analysis method that combines the bidirectional gated recurrent unit neural network (BiGRU) with the attention mechanism for Weibo topic sentiment analysis[2]. Their method achieved promising results, outperforming traditional neural network models in terms of accuracy. However, the authors acknowledged limitations in handling complex semantics like irony and lack of fine-grained sentiment analysis, indicating areas for future research.

Further, Lyu et al. (2020) explored public attitudes towards child abuse in mainland China by applying sentiment analysis to Weibo comments[3]. This study shed light on the emotional vocabulary and keywords related to resentment and vengefulness associated with child abuse. Nevertheless, this sentiment analysis was context-specific and did not extend to a broader range of topics.

Yang et al. (2019) demonstrated the use of extended vocabulary and CNNs for sentiment analysis of Weibo comment texts[4]. This research highlighted the efficiency of CNNs in handling large data sets and their high accuracy in sentiment classification.

Another work by Jia and Peng (2022) utilized a BiLSTM model with an attention mechanism for sentiment analysis regarding the Double Reduction Policy on Weibo[5]. Their method unveiled key themes related to the policy and successfully traced the online public sentiment trends.

Li et al. (2021) proposed a novel model combining BERT and deep learning for Weibo text sentiment analysis, demonstrating substantial improvements over similar models[6]. This research exemplified the integration of advanced NLP models with deep learning for sentiment analysis tasks.

Lastly, Chen (2015) emphasized the efficacy of CNNs for sentence classification, thus substantiating their potential for sentiment analysis[7]. The study underscored the superior performance of deep CNNs over traditional methods, highlighting their capability to capture intricate semantic features.

In summary, while different machine learning and deep learning techniques have shown success in Weibo sentiment analysis, CNNs have shown significant promise due to their hierarchical feature learning and efficient handling of high-dimensional data. However, the effectiveness of CNNs for sentiment analysis on a broader range of Weibo data still requires extensive exploration. This present study aims to address this gap by investigating the application of CNNs for sentiment analysis on Weibo data.

### III. PROBLEM DEFINITION

Despite the substantial progress in sentiment analysis using machine learning and deep learning techniques, there remains a significant challenge in understanding and interpreting the sentiments in Weibo data. This challenge primarily stems from the nature of the Weibo platform and the Chinese language. As one of the most popular microblogging sites in China, Weibo contains a massive volume of user-generated content, which is predominantly in Chinese. The Chinese language, characterized by its rich set of homonyms, extensive use of idioms, and lack of space-separated words, poses unique obstacles to the task of sentiment analysis.

The primary problem that this study aims to address is the enhancement of sentiment analysis performance on Weibo data by leveraging the power of Convolutional Neural Networks (CNNs). Previous studies have highlighted the capability of CNNs in processing high-dimensional data and extracting intricate hierarchical features, which are especially relevant for sentiment analysis on Weibo data. The application of CNNs allows us to capture local dependencies in the data and construct meaningful, higher-level representations of input sentences, which could improve the accuracy of sentiment prediction.

However, the application of CNNs for sentiment analysis on Weibo data remains largely unexplored, and its effectiveness across a broad range of topics on Weibo is still uncertain. Therefore, this study is motivated by the following research question: Can CNNs effectively enhance the performance of sentiment analysis on Weibo data across various topics?

Our study intends to investigate the suitability and efficiency of CNNs for sentiment analysis on Weibo data. We hypothesize that, given their demonstrated strength in other areas of natural language processing, CNNs could significantly boost the performance of sentiment analysis on Weibo, thus providing more accurate and granular insights into public opinion as reflected on this platform.

### IV. METHODOLOGY

This section details the methodology used in our experiment, which aimed to classify sentiments of Weibo posts into three categories: positive, neutral, and negative.

#### A. Data Collection

The dataset for this study was obtained from Baidu's PaddlePaddle Artificial Intelligence platform, specifically the dataset titled "weibo_senti_100k". This dataset comprises 119,988 Weibo posts, each labeled as either positive (1) or negative (0) according to the sentiment conveyed.

The distribution of sentiment categories in the original dataset was analyzed and visualized using a bar plot. The analysis revealed an almost balanced distribution of positive and negative sentiments. Specifically, there were 59,995 instances of negative sentiments and 59,993 instances of positive sentiments.

By plotting these sentiment counts, we were able to visually confirm the equal distribution of sentiment categories. This equal distribution is crucial as it can potentially reduce bias in the machine learning model we intend to train.

Fig. 1. Number of sentiment categories in original dataset

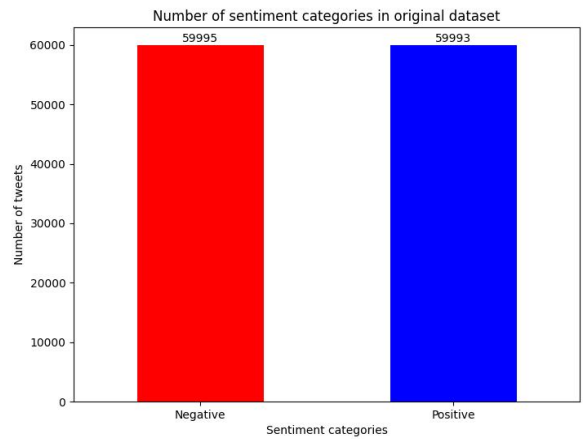

#### B. Data Preprocessing

The data preprocessing stage was crucial to ensure that the raw Weibo post data was converted into a format suitable for feeding into the Convolutional Neural Network (CNN) model.

The data preprocessing started with the removal of user mentions from the Weibo posts. This was done by replacing any text that started with '@' and ended with a whitespace character. Following this, all punctuation was stripped from the Weibo posts.

Next, the data was tokenized using Jieba, a popular Chinese text segmentation tool. This converted the continuous text into discrete tokens, or words, which were later used as input for the CNN model.

Subsequently, a list of Chinese stop words was loaded, and these stop words were removed from the tokenized Weibo posts. Stop words are high-frequency words that often carry little meaningful information, such as 'the', 'and', 'is', etc., and are commonly removed in text preprocessing to decrease the dimensionality of the data and to focus on important words.

This preprocessing resulted in some empty reviews, which were then identified and removed from the dataset, resulting in a final count of 117,282 reviews.

The sentiment of the cleaned and tokenized Weibo posts was then determined using the SnowNLP library. A sentiment score between 0 (negative) and 1 (positive) was assigned to each post. These sentiment scores were then classified into three categories: negative (sentiment score less than 0.3), neutral (sentiment score between 0.3 and 0.7), and positive (sentiment score higher than 0.7).

The data was visualized again using a bar plot to understand the distribution of the sentiment categories after preprocessing. The plot revealed an imbalance in sentiment categories, with positive sentiments being the most common (72,909 instances), followed by negative sentiments (26,368 instances) and neutral sentiments (18,005 instances).

The processed data was then saved into a new CSV file for further use in the study.

Fig. 2. Number of sentiment categories

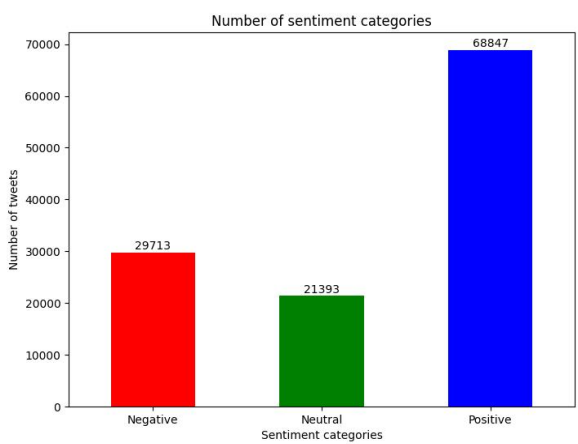

*C. Dataset Balancing*

Balancing the dataset is an essential step in training a machine learning model, particularly when dealing with multi-class classification tasks. In the case of this study, the original binary classification dataset was transformed into a three-class dataset during preprocessing, where positive sentiment was significantly more common than either neutral or negative sentiment. This imbalance could have resulted in a model biased towards predicting positive sentiment, leading to a decrease in performance.

To address this, the study employed an oversampling technique using the RandomOverSampler function from the imblearn library. This function works by randomly duplicating instances from the minority classes until all classes have the same number of instances. Prior to oversampling, the data was split into a training set (80% of the data) and a test set (20% of the data), using the train_test_split function from the sklearn library, to prevent data leakage and ensure a fair evaluation of the model's performance.

The oversampling was applied only to the training set, resulting in a balanced distribution of 14,434 instances for each sentiment category (negative, neutral, positive). The balanced training set was then ready to be used to train the CNN model.

A bar plot of the sentiment category distribution after oversampling confirmed the success of the balancing process, with all categories represented equally. Ensuring a balanced training set is crucial as it allows the model to learn from an equal number of instances from each class, avoiding bias towards the majority class and improving the model's ability to generalize to unseen data.

Fig. 3. Number of sentiment categories after oversampling balance

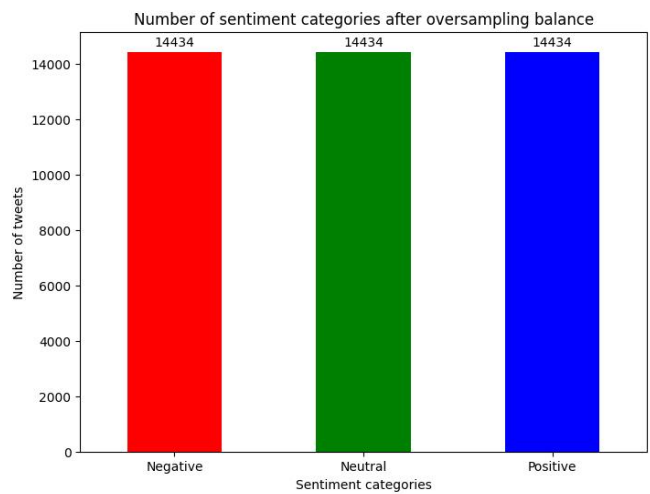

*D. Feature Extraction*

Feature extraction is a crucial step in machine learning and deep learning pipelines. It involves transforming raw data into a format that is compatible with the learning algorithms. In this study, we use a Convolutional Neural Network (CNN) model, which requires a specific format for its input.

The study employs the Tokenizer utility from the Keras library to tokenize the preprocessed text data, converting each word into an integer. The parameter 'num_words' is set to 5000, meaning the Tokenizer will only consider the top 5000 most frequent words in the corpus, ignoring rare words. This approach can help reduce the dimensionality of the input data and speed up computation without sacrificing too much information, as rare words typically contribute less to the overall understanding of a text.

The sequences of integers obtained after tokenization then need to be padded to ensure that they all have the same length, which is a requirement for feeding data into a CNN. This is done using the pad_sequences utility from the Keras library, with the maximum sequence length ('maxlen') set to 400.

*E. Model Architecture*

In this study, we adopt a Convolutional Neural Network (CNN) model to perform sentiment classification. The choice of CNN is made based on a series of experimental iterations. Initially, simpler models were tried but resulted in relatively low evaluation scores, indicating inadequate learning capacity. More complex models, like transformers, were considered; however, they required extensive computational resources and took over 60 hours to train, which was infeasible given our computational constraints. Therefore, CNNs strike a balance by providing enough complexity to capture the intricate patterns in text data while remaining relatively lightweight and efficient to train.

The constructed CNN model consists of several layers:

*1) Embedding Layer:*

This layer transforms the integer-encoded vocabulary into a dense vector representation. It takes 5000 (the size of the vocabulary) as the input dimension and outputs 50-dimensional vectors.

*2) Dropout Layer:*

A dropout layer is used right after the embedding layer to prevent overfitting. It randomly sets 20% of the input units to 0 at each update during training time.

*3) Convolutional Layer (Conv1D):*

This layer applies 250 filters, each of size 3 (kernel size) to the input data. It uses the 'relu' activation function and a stride of 1.

*4) GlobalMaxPooling1D Layer:*

This layer reduces the output of the convolutional layer by taking the maximum value over the time dimension for each feature map.

*5) Dense (Fully Connected) Layer:*

This layer has 250 neurons and uses the 'relu' activation function. It connects each input to each output within its layer.

*6) Dropout Layer:*

Another dropout layer is used to prevent overfitting, randomly setting 20% of the input units to 0 at each update during training time.

*7) Output Layer:*

The final dense layer uses the 'softmax' activation function and has 3 neurons, corresponding to the three sentiment categories.

The model is compiled using the 'categorical_crossentropy' loss function and 'adam' optimizer, and it is evaluated based on accuracy. The training is carried out for five epochs, with an early stopping strategy monitoring the validation loss and patience of 2.

Fig. 4. CNN model structure diagram

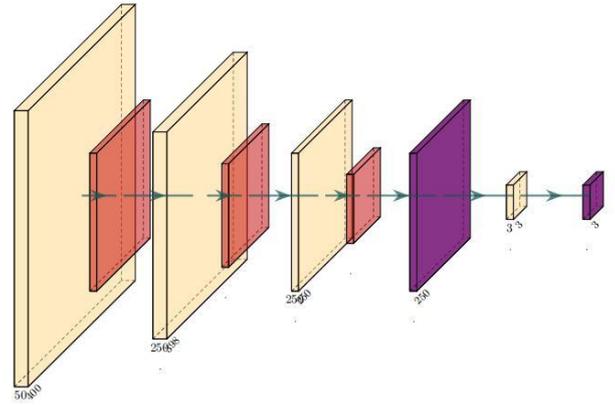

After training, the model predicts sentiment labels for the test set. The model performance is evaluated by precision, recall, and F1-score, with results indicating reasonable performance across all three sentiment classes.

TABLE I.　THE MODEL STRUCTURE IS AS FOLLOWS: FIGURES AND TABLES

| Layer (type) | Output Shape | Param # |
|---|---|---|
| embedding (Embedding) | (None, 400, 50) | 250000 |
| dropout (Dropout) | (None, 400, 50) | 0 |
| conv1d (Conv1D) | (None, 398, 250) | 37750 |
| global_max_pooling1d (GlobalMaxPooling1D) | (None, 250) | 0 |
| dense (Dense) | (None, 250) | 62750 |
| dropout_1 (Dropout) | (None, 250) | 0 |
| activation (Activation) | (None, 250) | 0 |
| dense_1 (Dense) | (None, 3) | 753 |
| activation_1 (Activation) | (None, 3) | 0 |

a. Note: The total parameters in the model are 346,253, and all of them are trainable.

V. RESULTS

Our sentiment analysis model, built on a Convolutional Neural Network (CNN), was trained and tested on the preprocessed and balanced dataset. We used accuracy as a training metric, with the model showing a steady increase in accuracy over the epochs. After the 3rd epoch, the validation loss started to increase, indicating the model may start overfitting the training data. Thanks to the early stopping callback, our training halted, preventing overfitting.

Upon evaluating the model on the test dataset, we achieved a macro-average F1-score of approximately 0.73. This metric provides a better measure of the incorrectly classified cases than the accuracy metric, as it takes both false positives and false negatives into account. The weighted average F1-score, precision, and recall are all approximately 0.73, showing balanced performance across the three sentiment classes (-1, 0, 1).

The performance metrics for each class in the test set are as follows:

TABLE II.　EXPERIMENTAL RESULTS OF CNN MODEL

| Sentiment Class | Precision | Recall | F1-Score | Support |
|---|---|---|---|---|
| -1 | 0.75 | 0.77 | 0.76 | 3637 |
| 0 | 0.62 | 0.68 | 0.65 | 3597 |
| 1 | 0.83 | 0.72 | 0.77 | 3566 |

b. Note: Sentiment classes are denoted as -1 for negative, 0 for neutral, and 1 for positive.

## VI. Discussion

Our Convolutional Neural Network (CNN) model delivered stable and reliable performance for the sentiment analysis task across multiple experiments. The success of this model can be attributed to several factors:

Firstly, the CNN architecture is inherently suitable for text data. While traditionally used for image analysis, CNNs can also be used for text classification by treating text as a form of one-dimensional sequence data. The ability of CNNs to automatically learn and extract valuable features from the text helped significantly in the task of understanding and classifying sentiment.

Secondly, we leveraged word embeddings as a form of feature extraction, transforming text data into numeric vectors that machines can understand. This step effectively captures the semantic relationships between words, providing rich representation for our model to learn from.

Lastly, our preprocessing steps, such as tokenizing, padding, and transforming labels into categorical format, prepared our text data in a way that made it easier for our model to learn and make accurate predictions.

Our results are comparable to many recent studies in sentiment analysis, demonstrating the efficacy of CNNs for this task. However, as compared to complex architectures like transformers, our model offered a balance between computational efficiency and predictive performance, making it a viable option when computational resources or time are limited.

Although the model's performance was reasonably good, it was noticed that the performance for neutral sentiment was slightly lower compared to positive and negative sentiments. This could be due to the inherent complexity in classifying neutral sentiments which often lack clear sentiment indicators.

The outcomes of this research could have a significant impact on various real-world applications, particularly those involving the analysis of public opinion, such as in social media analysis, market research, and political studies. Improved sentiment analysis can help businesses understand their customer feedback better, provide personalized recommendations, or help policymakers understand public opinion on certain issues. Future work can include exploring different architectures, like Recurrent Neural Networks (RNN) or transformers, or utilizing more complex pre-trained models like BERT, to further improve the model's understanding of the nuances in sentiment.

## VII. Conclusion

Our study aimed to address the challenging task of sentiment analysis on a dataset derived from Weibo. Through the application of a Convolutional Neural Network (CNN), we aimed to build a model that could effectively understand and categorize sentiments expressed in tweets.

Our methodology involved preprocessing the text data, transforming it into numeric vectors using word embeddings, and feeding this into our CNN model. This model was then trained and evaluated using various metrics such as accuracy, precision, recall, and F1-score. The results demonstrated that our CNN model provided a balanced performance across the three sentiment categories - positive, neutral, and negative.

This research underlines the viability of using CNNs for sentiment analysis tasks and has implications for various real-world applications. The capacity to understand and categorize public sentiment from social media can provide valuable insights for businesses, marketers, and policymakers.

Looking ahead, there are several exciting directions for future research. The model's performance could potentially be improved by experimenting with different architectures, such as Recurrent Neural Networks or transformers. Another promising direction could involve leveraging more complex pre-trained models like BERT, which could improve the model's understanding of context and nuances in language. Lastly, while our study focused on Weibo Chinese data, it would be interesting to test the model on different social media platforms, which may present unique linguistic features and challenges for sentiment analysis.